\pdfoutput=1

\documentclass[11pt]{article}

\usepackage{acl}

\usepackage{times}
\usepackage{latexsym}
\usepackage{amsmath}
\usepackage{amssymb}
\usepackage{multirow}

\usepackage[T1]{fontenc}

\usepackage[utf8]{inputenc}

\usepackage{microtype}

\title{A Neural Pairwise Ranking Model for Readability Assessment}

\author{Justin Lee\thanks{Work done during an internship at National Research Council, Canada}\\
  University of Toronto \\
  Toronto, Canada \\
  \small{\texttt{chunhin.lee@mail.utoronto.ca}} \\\And
  Sowmya Vajjala \\
National Research Council \\
Ottawa, Canada \\
  \small{\texttt{sowmya.vajjala@nrc-cnrc.gc.ca}} \\}

\begin{document}
\maketitle
\begin{abstract} 
Automatic Readability Assessment (ARA), the task of assigning a reading level to a text, is traditionally treated as a classification problem in NLP research. In this paper, we propose the first neural, pairwise ranking approach to ARA and compare it with existing classification, regression, and (non-neural) ranking methods. We establish the performance of our model by conducting experiments with three English, one French and one Spanish datasets. We demonstrate that our approach performs well in monolingual single/cross corpus testing scenarios and achieves a zero-shot cross-lingual ranking accuracy of over 80\% for both French and Spanish when trained on English data. Additionally, we also release a new parallel bilingual readability dataset in English and French. To our knowledge, this paper proposes the first neural pairwise ranking model for ARA, and shows the first results of cross-lingual, zero-shot evaluation of ARA with neural models. 
\end{abstract}

\section{Introduction}
Automatic Readability Assessment is the task of assigning a reading level for a given text. It is useful in many applications from selecting age appropriate texts in classrooms \cite{Sheehan.Kostin.ea-14} to assessment of patient education materials \cite{Sare.Patel.ea-20} and clinical informed consent forms \cite{Perni.Rooney.ea-19}. Contemporary NLP approaches treat it primarily as a classification problem which makes it non-transferable to situations where the reading level scale is different from the model. Applying learning to rank methods has been seen as a potential solution to this problem in the past. Ranking texts by readability is also useful in a range of application scenarios, from ranking search results based on readability \cite{Kim.Collins-Thompson.ea-12,Fourney.Ringel.ea-18} to controlling the reading level of machine translation output \cite{Agrawal.Carpuat-19,Marchisio.Guo.ea-19}. However, exploration of ranking methods has not been a prominent direction for ARA research. Further, recent developments in neural ranking approaches haven't been explored for this task yet, to our knowledge.

ARA typically relies on the presence of large amounts of data labeled by reading level. Further, although linguistic features are common in ARA research, it is challenging to calculate them for several languages, due to lack of available software support. Though there is a lot of recent interest in neural network based cross-lingual transfer learning approaches for various NLP tasks, there hasn't been much research in this direction for ARA yet. 

With this context, we address two research questions in this paper:
\begin{enumerate}\itemsep-1mm
 \item Is neural, pairwise ranking a better approach than classification or regression for ARA?
    \item Is zero-shot, cross-lingual transfer possible for ARA models through ranking? 
\end{enumerate}

The main contributions of this paper are:
\begin{enumerate}\itemsep-1mm
    \item A new neural pairwise ranking model with an application to automatic readability assessment.
    \item Demonstration of the use pairwise ranking to achieve cross-corpus compatibility in ARA.
    \item First evidence of zero shot, neural cross-lingual transfer in ARA.
    \item A new parallel readability dataset, \textit{Vikidia En/Fr}, the first of its kind in ARA research. 
\end{enumerate}

The rest of this paper is organized as follows: Section~\ref{sec:relw} gives an overview of related research and Section~\ref{sec:model} describes the proposed neural pairwise ranking model. The next two Sections (~\ref{sec:setup} ~\ref{sec:results}) describe our experimental setup and discuss the results. Section~\ref{sec:disc} summarizes our findings and discusses the limitations of this approach. 

\section{Related Work}
\label{sec:relw}
Readability Assessment has been an active area in educational research for almost a century. Early research on this topic focused on the creation of readability "formulae", which relied on easy to calculate measures such as word and sentence length, and presence of words from some standard word list \cite{Lively.Pressey-23,Flesch-48,Stenner-96}. More than 200 such formulae were proposed in the past few decades \newcite{Dubay-07}. The advent of NLP and machine learning resulted in more data driven research on ARA over the past two decades. Starting from statistical language models \cite{Si.Callan-01}, a range of lexical and syntactic features \cite{Heilman.Collins-Thompson.ea-07,Petersen.Ostendorf-09,Ambati.Reddy.ea-16} as well as inter-sentential features \cite{Pitler.Nenkova-08,Todirascu.Francois.ea-13,Xia.Kochmar.ea-16} were developed in the past. Features motivated by related disciplines such as  psycholinguistics \cite{Howcroft.Demberg-17}, second language acquisition \cite{Vajjala.Meurers-12} and cognitive science \cite{Feng.Elhadad-09} were also explored for this task. 

In the past few years, ARA research has been primarily focused on textual embeddings and deep learning based architectures. Word embeddings in combination with other attributes such as domain knowledge or language modeling \cite{Cha.Gwon.ea-17,Jiang.Gu.ea-18} and a range of neural architectures, from multi attentive RNN \cite{Azpiazu.Pera-19} to deep reinforcement learning \cite{Mohammadi.Khasteh.ea-19} were proposed. Recent research explored combining transformers with linguistic features \cite{Deutsch.Jasbi.ea-20,Meng.Chen.ea-20,Lee.Jang.ea-21,Imperial-21}. 

Although a lot of this research evolved on English, the past decade saw ARA research in other languages such as German \cite{Hancke.Vajjala.ea-12}, French \cite{Francois.Fairon-12}, Italian \cite{DellOrletta.Montemagni.ea-11}, Bangla \cite{Sinha.Sharma.ea-12} etc., which employed language specific feature sets. While most ARA research modeled one language at a time, some research created language agnostic feature sets and architectures and experimented with 2 to 7 languages \cite{Shen.Williams.ea-13,Azpiazu.Pera-19,Madrazo.Pera-20a,Madrazo.Pera-20b,Martinc.Pollak.ea-21,Weiss.Chen.ea-21}. Although only one language is considered per model in all this research, there are two important exceptions. \newcite{Madrazo.Pera-20b} explored whether combining texts from related languages during training improves ARA performance for low resource languages. \newcite{Weiss.Chen.ea-21} used a model trained on English texts on German, based on a common, broad set of handcrafted linguistic features. However, to our knowledge, zero-shot cross lingual transfer of neural network architecture based approaches, without any handcrafted features, was not explored for this task in the past. 

ARA is traditionally treated as a classification problem in NLP research, although there are some exceptions. \newcite{Heilman.Collins-Thompson.ea-08} compared linear, ordinal, and logistic regression and concluded that ordinal regression with a combination of lexical and grammatical features worked the best for ARA, although classification approaches still dominated subsequent research on the topic. There is some past work that considered ARA as a pairwise ranking problem, using SVM/SVM$^{rank}$ and hand crafted linguistic features \cite{Pitler.Nenkova-08,Tanaka-Ishii.Tezuka.ea-10,Ma.Fosler-Lussier.ea-12,Mesgar.Strube-15,Ambati.Reddy.ea-16,Howcroft.Demberg-17}. While \newcite{Tanaka-Ishii.Tezuka.ea-10} and \newcite{Ma.Fosler-Lussier.ea-12} showed that ranking performs better than traditional features and classification/regression respectively, \newcite{Xia.Kochmar.ea-16} did not find ranking to be consistently better across the board. Given this background, we take a fresh look at the application of ranking for ARA, by proposing a new neural pairwise ranking model. 

\section{Neural Pairwise Ranking Model}
\label{sec:model}
The data for our pairwise ranking model takes the form of (document, reading level) pairs. Let $X = [(x_1, y_1),...,(x_n, y_n)]$ be $n$ such pairs, where $x_i$ is the vector representation for document $i$ and $y_i$ is the corresponding reading level. We then construct $m$ pairwise permutations from $X$ to form $X'$. The members of $X'$ are constructed as follows: if a pair of documents and reading levels $(x_i, y_i)$ and $(x_j, y_j)$ are chosen, then both permutations $((x_i, x_j), (y_i, y_j))$ and $((x_j, x_i), (y_j, y_i))$ are added to $X'$.  

The neural pairwise ranking model (\textit{$NPRM$}) aims to maximize
\begin{align*}
    P(y_i > y_j|x_i, x_j)
\end{align*}

Formally, this is parametrized as 
\begin{align*}
    P(y_i > y_j|x_i, x_j) & \triangleq NPRM(x_i, x_j)\\
    & = softmax(\psi(f(x_i, x_j))) \\
    & = [s_{ij1}, s_{ij2}]
\end{align*}

where $f$ is a neural model, $\psi$ is a flexible function, $s_{ij1}$ represents the predicted score of $P(y_i > y_j|x_i, x_j)$ and $s_{ij2}$ represents the predicted score of 1 - $P(y_i > y_j|x_i, x_j)$.  
Training labels are created as 
\begin{align*}
    y'_{ij} = \begin{cases} 
                [1, 0] & \text{ if } y_i\geq y_j \\
                [0, 1] & \text{ if } y_i < y_j  \\
            \end{cases}
\end{align*}

We then calculate the loss function as 
\begin{align*}
    L = - y'_{ij1} \cdot log(s_{ijk1}) - y'_{ij2} \cdot log(s_{ijk2})
\end{align*}
and back-propagate the errors with stochastic gradient descent. This loss function is known as the Pairwise Logistic Loss \cite{Han.Wang.ea-20}.

\paragraph{Implementation} 
Our neural pairwise ranking model ($NPRM$) consists of a BERT \cite{devlin2018bert} model as $f$ and a fully connected layer as $\psi$. We evaluate the performance of the pairwise ranking approach as follows: for a list of texts to be ranked of size $S$ and each text $x_a$ within the list, $1 \leq a \leq S$  we compute 
\begin{align*}
   Score(x_a) =  \sum_{b\neq a}^{}NPRM(x_a, x_b)
\end{align*}
We then rank each text $x_a$ by $Score(x_a)$ in descending order.

This pairwise ranking framework allows for $NPRM$ to model relative reading difficulties between texts. While other neural methods have been proposed with more sophisticated learning objectives for ranking problems in the past \cite{Wang.Li.ea-18, Ai.Wang.ea-19}, these methods require fixed-size inputs to rank. The $NPRM$ only needs a minimum of two texts to form a ranking for each, and the aggregation process of scores between pairwise permutations of texts can easily produce rankings for an arbitrary list size larger than two. The  aggregation process also produces a bounded (by the list size), but continuous score for each document, which results in a ranking with no ties, as long as the input documents are different. 

Additionally, the choice of $f$ in the $NPRM$ framework can allow for multi-lingual predictability, improved performance, or improved efficiency. Due to the flexible modeling structure that the $NPRM$ maintains, we hypothesize that with the aid of a multilingual language model, zero-shot cross lingual ARA assessment may also be possible with $NPRM$. We demonstrate this possibility later in the paper\footnote{All code for the model and experiments is at: \url{https://github.com/jlee118/NPRM/}}. 

\section{Experimental Setup}
\label{sec:setup}
We describe our experimental setup in terms of the datasets used, modeling and evaluation procedures, in this section. 
\subsection{Datasets}
We experimented with three English, one Spanish and one French datasets, which are described below. All the datasets contain texts in multiple reading level versions (in a given language). We call such grouping of a given text in multiple reading levels a \textit{slug}. We used the first two English datasets for training and testing our models, and the remaining three datasets only as test sets.

\paragraph{NewsEla-English (NewsEla-En):} NewsEla\footnote{NewsEla corpus can be requested from: \url{https://NewsEla.com/data/}} provides leveled reading content, which is aligned with the common core educational standards \cite{Porter.McMaken.ea-11}, and contains texts covering grade 2 to grade 12. It follows the Lexile \cite{Stenner-96} framework to create such leveled texts. It was first used in NLP research by \newcite{Xu.Callison-Burch.ea-15} and has been a commonly used corpus for ARA and text simplification in the recent past. The English subset of the NewsEla dataset contains 9565 texts distributed across 1911 slugs. Slugs may or may not contain texts for the full range of reading levels available i.e., each text does not have all reading level versions.

\paragraph{OneStopEnglish (OSE): } This consists of articles sourced from The Guardian newspaper, rewritten by teachers into three reading levels (beginner, intermediate, advanced) \cite{Vajjala.Lucic-18} and has been used as a bench marking dataset for ARA in the recent past. This dataset contains 189 slugs and 3 reading levels, summing to a total of 567 texts (each slug has one text in three versions).

\paragraph{NewsEla-Spanish (NewsEla-Es): } This is the Spanish subset within the existing NewsEla dataset and contains 1221 texts distributed across 243 slugs and 10 reading levels. Similar to NewsEla-En, each slug does not have all 10 levels in it. 

\paragraph{Vikidia-En/Fr: } Vikidia.org\footnote{\url{https://www.vikidia.org/}} is a children's encyclopedia, with content targeting 8-13 year old children, in several European languages. Our dataset contains 24660 texts distributed across 6165 slugs and 2 reading levels, for English (\textit{Vikidia-En}) and French (\textit{Vikidia-Fr}) respectively i.e., each text in the corpus has four versions: en, en-simple, fr and fr-simple, and there are 6165 slugs in total. \newcite{Azpiazu.Pera-19}'s experiments used data from this source. However, the data itself is not publicly available. The uniqueness of the current dataset is that these are parallel, document level aligned texts in four versions - en, en-simple, fr, fr-simple. While we did not create paragraph/sentence level alignments on the corpus, we hope that this will be a useful dataset for future English and French research on ARA and Automatic Text Simplification. This is the first such dataset in ARA, and perhaps the first readily available French readability dataset. It can be accessed at: \url{https://zenodo.org/record/6327828}

\subsection{Classification, Regression and Ranking models}
Our primary focus in this paper is on the pairwise ranking model. However, we also compared the performance of other classification, regression, and ranking approaches with our pairwise ranking model to establish strong points of comparison.

\paragraph{Feature representation: } While the use of linguistic features, and more recently, contexual embeddings, have been explored in ARA, non-contextual embeddings were not explored much. Hence, in this paper, we employ three non-contextual embeddings (GloVe \cite{Pennington.Socher.ea-14}, Word2vec \cite{mikolov2013efficient}, fastText \cite{Bojanoswki.Grave.ea-17}) for training classification/regression/ranking models.  Document-level embeddings are obtained by aggregating and averaging word-level embeddings for each token in the text. In addition, we also used a BERT \cite{devlin2018bert} based classifier.

\paragraph{Classification}
The following models were used for formulating baselines and comparisons for classification. Reading levels are treated as class labels, and evaluation is done via 5-Fold cross validation.  
\begin{itemize}
    \item Non-contextual embeddings fed into an SVM \cite{Boser.Guyon.ea-92} classifier 
    \item Non-contextual embeddings fed into a Hierarchical Attention Network (HAN) \cite{yang2016hierarchical}. This model was used with and without linguistic features in the past, for reading level classification (\newcite{Deutsch.Jasbi.ea-20} and \newcite{Martinc.Pollak.ea-21} respectively).  
    \item 110-M parameter, 12-layer, BERT model with a fully connected layer and a softmax output. The model is then fine-tuned on the classification task with categorical cross-entropy loss.
\end{itemize}
 
\paragraph{Regression}
The following models were used for formulating baselines for regression. Reading levels are treated as continuous outputs, and results are obtained through 5-Fold cross validation. 
\begin{itemize}
    \item Non-contextual word-level embeddings as input into an Ordinary Linear Regression (OLS) model. 
    \item 110-M parameter  BERT  model with a fully connected layer.  The model is then fine-tuned on the regression task with the mean squared error loss and will be referred to as \textit{regBERT} in this paper. 
\end{itemize}

\paragraph{(non-neural) Pairwise Ranking}
We employ an SVMRank model with a pairwise ranking framework similar to $NPRM$, but using the non-contextual word embeddings for feature extraction. Input features for the SVM model are obtained by differencing the obtained embeddings in the following manner: for any text representations $x_i, x_j$, with reading levels $y_i, y_j$, form training examples as $x'_i = x_i - x_j$ and $x'_j = x_j - x_i$,  and training labels as: 

{\centering
  $ \displaystyle
\begin{aligned}
    y'_i = \begin{cases} 
                1 & y_i\geq y_j \\
                0 & y_i < y_j  \\
            \end{cases}\\
    y'_j = \begin{cases} 
                1 & y_j\geq y_i \\
                0 & y_j < y_i  \\
            \end{cases}\\
\end{aligned}
 $ 
\par}
Predicted scores are aggregated in the same manner as in $NPRM$ to form rankings.  Results are obtained through 5-Fold cross validation.

\subsection{Pairwise Ranking Training}
To control for the variation in the text introduced by different topical content, the training and prediction process for the SVMRank and $NPRM$ aggregates the text by their slug designations before forming pairwise permutations. As a result, the pairwise permutations are constructed from the text within a slug. Note that slug is used for training and testing the model, but isn't really required while using the model for prediction. The trained model only takes a list of texts as inputs and returns a ranked list based on readability.

For controlling the computation time, we fixed the number of pairwise comparisons per slug ($m$ in NPRM) to 3 levels. i.e., In datasets with more than 3 levels per slug (NewsEla-En and NewsEla-Es) we choose texts with the highest and lowest reading levels within a slug, and sample the third text from a reading level in between. Note that this will not affect the ability of the model to rank a list of texts where $m$ is higher than 3. As with all baselines, results from $NPRM$ are obtained through 5-Fold cross validation.

\subsection{Evaluation}
Accuracy and F1-score are reported for classification and mean-absolute error (MAE) and mean-squared error (MSE) are reported for regression. To evaluate ranking performance, we calculate the Normalized Discounted Cumulative Gain (NDCG), Spearman's Rank Correlation(SRC), Kendall's Tau Correlation Coefficient (KTCC), and the percentage of slugs ranked completely correct, which we denote as Ranking Accuracy (RA). There is some work on evaluating ranking in NLP \cite{Lapata-06,Katerenchuk.Rosenberg-16}, without any consensus on the most suited metric. Hence, we chose to report multiple metrics instead of one, based on the commonly reported measures for such tasks.

We compare classification and regression predictions too using ranking metrics, in addition to traditional measures. To examine the ranking performance, the texts from each dataset are first grouped by their slugs. Then, ground-truth ranking of the texts within the slugs are compared against the rankings formed from the predicted scores of the models. For NDCG, we used the ground-truth reading levels as the relevance score. For all the metrics, We took the model predictions as is, and did not employ specific means to address ties (which can happen in classification). The metrics themselves address ties in different ways. NDCG averages ties in predicted scores, KTCC penalizes ties in ground truth and predicted scores, and SRC calculates the average rank of ties. Ranking accuracy does not handle ties.  

\subsection{Statistical Significance Testing}
We used Wilcoxon's signed rank test \cite{wilcoxon}, a non-parametric statistical hypothesis test to examine whether the performance differences between $NPRM$ and other methods are statistically significant, when the metrics are close to each other. Ranking metrics per slug from a sample per model are aggregated, and are then compared for any two models. 

\subsection{Technical Implementation}
Non-neural machine methods used the sklearn \cite{Pedregosa.Varoquaux.ea-11} library. The HAN model is a Keras implementation\footnote{\url{https://github.com/tomcataa/HAN-keras}}. Transformers library \cite{Wolf.Debut.ea-20} was used for accessing and fine-tuning BERT and mBERT based models (\textit{bert-base-uncased} and \textit{bert-base-multilingual-uncased} models were used).  TF-Ranking library \footnote{\url{https://github.com/tensorflow/ranking}} \cite{Pasumarthi.Bruch.ea-19} was used for accessing the Keras-compatible Pairwise Logistic Loss function. SciPy\cite{Virtanen.Gommers.ea-20} was used for statistical significance testing. 

The Word2vec embeddings are pre-trained on English Google News \cite{Mikolov.Chen.ea-13}. The fasttext embeddings contain 1-million word vectors and are trained on subword information from Wikipedia 2017 \cite{Bojanoswki.Grave.ea-17}. The GloVe embeddings are trained on the Wikipedia 2014 and Gigaword 5 corpus \cite{Pennington.Socher.ea-14}. All three are accessed through gensim\footnote{\url{https://radimrehurek.com/gensim/}}.  

\section{Results}
\label{sec:results} 
We performed within corpus evaluation for classification, and within/cross corpus as well as cross-lingual evaluation for regression and ranking. We did not employ classification approaches in the last two evaluation settings as there is no way of resolving ties with classifier predictions. Further, regression and ranking gave better performance than classification in monolingual, within-corpus settings.

\subsection{Classification}
We trained models using \textit{Newsela-En} and \textit{OSE} datasets respectively in a five fold CV setup, for classification. Table~\ref{tab:justclass} shows the performance of our best performing model in terms of traditional classification metrics, comparing with the state of the art. 

\begin{table}[htb]
\begin{tabular}{|p{4.95cm}|c|}
\hline
Model & weighted-F1\\ \hline
\multicolumn{2}{|c|}{NewsEla-En} \\ \hline
HAN \cite{Martinc.Pollak.ea-21} & \textbf{0.81}\\
BERT & 0.74 \\ \hline
\multicolumn{2}{|c|}{OSE} \\ \hline
HAN \cite{Martinc.Pollak.ea-21} &0.79 \\
BERT & 0.93\\ 
BART \cite{Lewis.Liu.ea-20}+Linguistic features \cite{Lee.Jang.ea-21} & \textbf{0.97} \\ \hline
\end{tabular}
\caption{Weighted-F1 for classification}
\label{tab:justclass}
\end{table}
In terms of traditional classification metrics, our approach achieves a lower performance than \newcite{Martinc.Pollak.ea-21} for NewsEla-En corpus, but higher performance on the OSE corpus. A more recent paper by \newcite{Lee.Jang.ea-21} reported further improvement with OSE, with an extensive set of linguistic features. Table~\ref{tab:rankingclass} shows the performance of all models in terms of the ranking metrics.

\begin{table}[htb]
\begin{tabular}{ |p{2.45cm}|p{0.85cm}|p{0.8cm}|p{0.8cm}| p{0.8cm}| }
 \hline
Model & NDCG & SRC &KTCC & RA\\
 \hline
  \multicolumn{5}{|c|}{NewsEla-En} \\
  \hline
BERT   & \textbf{0.999}    & \textbf{0.992} &  \textbf{0.985} & 0.927 \\
GloVe + HAN & 0.991 & 0.985 & 0.971 & 0.971\\
GloVe + SVM & 0.947 & 0.866 & 0.796 & 0.981\\
fasttext + HAN & 0.991 & 0.985 & 0.972 & 0.971\\
fasttext + SVM & 0.996 & 0.939 & 0.892 & \textbf{0.990}\\
\hline
   \multicolumn{5}{|c|}{OSE} \\
 \hline
BERT&  0.963   & 0.808   & 0.808 & 0.825\\
GloVe + HAN & 0.938 & 0.741 & 0.741 & 0.841\\
Glove + SVM & 0.875 & \textbf{0.931} & \textbf{0.930} & \textbf{0.963}\\
fasttext + HAN & \textbf{0.964} & 0.857 & 0.854 & 0.899\\
fasttext + SVM & 0.867 & 0.763 & 0.763 & 0.884\\
 \hline
\end{tabular}
\caption{Ranking Metrics for Classification Evaluation}
\label{tab:rankingclass}
\end{table}
When evaluating the classification models in terms of ranking metrics, we notice some differences among the models evaluated using NewsEla-En and OSE. There is relatively less variation among different Newsela-En models for NDCG, compared to SRC, KTCC, and RA. We see larger variations across OSE models for all the metrics. It is interesting to note that the non-contextual embeddings perform competitively with BERT in terms of the ranking metrics and are all better than BERT in terms of ranking accuracy. Overall, The NewsEla-En + BERT classifier achieves the highest average NDCG, SRC, and KTCC, and the NewsEla + fasttext + SVM combination achieves the highest ranking accuracy. For OSE, the Glove+SVM classifier achieves the highest SRC, KTCC, and RA while fastText+HAN and BERT models achieve better scores in terms of NDCG. 

All the ranking metrics in general seem to have higher scores with NewsEla-En trained models, than OSE models. This could potentially be due to the larger dataset size, as well as the fact that NewsEla-En covers a broader reading level scale. Although the classification models seem to generally perform well on ranking metrics too, it has to be noted that there is no inherent means within classification to distinguish between ties, where the model predicts the same class for two documents of different reading levels. Hence, it is not feasible to continue to use classifiers as rankers. This evaluation is to be seen only means of comparing classification, regression, and ranking with a common set of metrics. 

\begin{table*}[htb!]
\centering
\begin{tabular}{|c|c|c|c|c|c|c|}
 \hline
 Model & MSE & MAE & NDCG & SRC & KTCC & RA\\
 \hline
 \multicolumn{7}{|c|}{NewsEla-En} \\
 \hline
 regBERT & \textbf{0.434}   & \textbf{0.460} & 0.999 & 0.997 & 0.994 & \textbf{0.977}\\
 gloVe+OLS & 2.310 & 1.212 &0.999 & 0.988 & 0.978 & 0.900\\
 word2Vec+OLS & 1.734    & 1.056& 0.999 & 0.996 & 0.992 & 0.961\\
 fasttext + OLS&  1.766   & 1.058& 0.999 & 0.997 & 0.994 & 0.971 \\
  \hline \multicolumn{7}{|c|}{OSE} \\
\hline
regBERT&  \textbf{0.260}   & \textbf{0.376} &0.986 & \textbf{0.944} & \textbf{0.929} & \textbf{0.905}\\
gloVe + OLS & 2.143 & 1.122& 0.989 & 0.857 & 0.834 & 0.794\\
 word2vec + OLS &  1.888   & 1.076 & 0.988 & 0.873 & 0.855 & 0.852 \\
 fasttext + OLS &  1.561   & 0.953 & \textbf{0.995} & 0.926 & 0.912 & 0.899\\ \hline
 \end{tabular}
\caption{Performance of Regression approaches}
\label{tab:regmetrics}
\end{table*}

\subsection{Regression}
Table~\ref{tab:regmetrics} shows the performance of all the regression models using both regression and ranking metrics. 

Although there are no other reported results of applying regression models on these datasets to our knowledge, the low MAE/MSE for both datasets indicate that regression models perform well for this problem. Like with classification, we notice that there is no huge difference among the contextual and non-contextual embeddings in terms of the ranking metrics. However, we notice some general differences between classification and regression approaches. In contrast to the classification models, when holding the training data and the regressor constant, models with GloVe embeddings perform worse than models using Word2vec or fasttext in regression specific metrics. When evaluating on ranking metrics, the regression models generally exhibit higher average NDCG, SRC and KTCC than the classification models. Again, like with classification evaluation, the differences across models in terms of the ranking metrics is larger for OSE compared to NewsEla-En. Overall, though, the neural regressor (regBERT) consistently performs better than the OLS regressor in terms of regression metrics, and is either comparable or better than OLS regressor in terms of all the ranking metrics. 

\subsection{Pairwise Ranking}

\begin{table}[htb]
\begin{tabular}{ |p{2.45cm}|p{0.85cm}|p{0.7cm}|p{0.8cm}| p{0.7cm}|}
 \hline
Model & Avg. NDCG & Avg. SRC & Avg. KTCC & RA\\
 \hline
\multicolumn{5}{|c|}{NewsEla-En} \\
\hline
NPRM BERT   & \textbf{0.999}    & 0.995 & 0.990 & 0.948\\
  \hline
word2vec + SVMRank & 0.997 & \textbf{0.997} & \textbf{0.997} & \textbf{0.979}\\
 \hline
fasttext + SVMRank & 0.998 & 0.995 & 0.991 & 0.957 \\
 \hline
GloVe + SVMRank  & 0.998 & 0.992 & 0.985 & 0.932 \\
 \hline 
 \multicolumn{5}{|c|}{OSE}\\
 \hline
NPRM BERT&  \textbf{0.997}  & \textbf{0.981} & \textbf{0.979} & \textbf{0.979}\\
 \hline
word2vec + SVMRank & 0.972 & 0.966 & 0.962 &0.958 \\
 \hline
fasttext + SVMRank & 0.991 & 0.947 & 0.940 & 0.931\\
 \hline
GloVe + SVMRank & 0.994 & 0.971 & 0.968 & 0.968\\
 \hline
\end{tabular}
\caption{Pairwise Ranking Evaluation}
\label{tab:rankingsingle}
\vspace{-4mm}
\end{table}
Table~\ref{tab:rankingsingle} shows the performance of pairwise ranking approaches on both the training datasets. When training on the NewsEla-En dataset, we observe that $NPRM$ outperforms at least one word-embedding + SVMRank combination in the ranking metrics, but only achieves the top score in NDCG when compared with word-embedding SVMRank methods. When training on the OSE dataset, $NPRM$ achieves the top score against the word-embedding + SVMRank combinations, but only NDCG was found to be statistically significant across all models.  Comparisons between $NPRM$ and the word-embedding + SVMRank combinations had p-values < 0.05 for NDCG.  For SRR, KTCC and RA, only the difference in scores between $NPRM$ and fasttext + SVMRank were found to be statistically significant. GloVe + SVMRank method produces the statistically equivalent scores in SRC, KTCC, and RA as $NPRM$. 

Overall, while there is no single approach that ranked as the best uniformly across all the three model settings (Tables ~\ref{tab:rankingclass}-~\ref{tab:rankingsingle}), BERT based models perform competitively with most of the ranking metrics. Table ~\ref{tab:allthreebert} presents a summary of the performance of BERT in classification, regression and ranking setups. 

\begin{table}[htb]
\begin{tabular}{ |p{2.5cm}|p{0.9cm}|p{0.7cm}|p{0.9cm}| p{0.8cm}|}
 \hline
Model & Avg. NDCG & Avg. SRC & Avg. KTCC & RA\\
 \hline
\multicolumn{5}{|c|}{NewsEla-En} \\
\hline 
BERT-Class. & 0.999 & 0.992 &  0.985 & 0.927\\
regBERT & 0.999 & 0.997 & 0.994 & 0.977\\
NPRM BERT & 0.999 & 0.995 & 0.990 & 0.948\\
 \hline 
 \multicolumn{5}{|c|}{OSE}\\
 \hline
BERT-Class. & 0.963   & 0.808   & 0.808 & 0.825\\
regBERT & 0.986 & 0.944 & 0.929 & 0.905\\
NPRM BERT & 0.997  & 0.981 & 0.979 & 0.979\\
 \hline
\end{tabular}
\caption{Classification vs Regression vs Ranking}
\label{tab:allthreebert}
\end{table}
For Newsela-En, all methods reported a high score of 0.999 for NDCG and regBERT is better with the other metrics. Testing for statistical significance between $NPRM$, regBERT and BERT classification showed that $NPRM$ is significantly better than BERT classifier (p$<$0.05) and there is no significant difference between $NPRM$ and regBERT. For OSE, NPRM BERT achieves a better performance for all metrics. We did not perform statistical significance testing in this case as the differences are larger.  

To conclude, when training and testing from the same distribution, regBERT and NPRM BERT perform better than BERT based classifier in terms of the ranking metrics. Since the performance is generally expected to degrade slightly in a cross-corpus setting compared to a within corpus evaluation, the rest of our experiments will only focus on regBERT and NPRM, and we don't report further experiments with a BERT classifier. 

\subsection{Cross-corpus Pair-wise Ranking}
In this experiment, we evaluated the performance of an ARA model trained with one English dataset, on other English datasets. Since NewsEla-en is the larger dataset with more diverse reading levels, we used that for training, and used OSE and Vikidia-En as test sets. Since regression scores can also be used to directly rank predictions, we compared the performance of NPRM with BERT based regression model. Table~\ref{tab:rankingcross} shows the results. 

\begin{table}[htb]
\begin{tabular}{ |p{2.45cm}|p{0.85cm}|p{0.7cm}|p{0.8cm}| p{0.7cm}|}
 \hline 
 \multicolumn{5}{|c|}{NPRM} \\ \hline 
 Testset & NDCG & SRC & KTCC & RA\\
 \hline
 OSE&  0.983 & 0.931 & 0.912 & 0.878\\ 
 \hline
 Vikidia-En & 0.991 & 0.950 & 0.950 & 0.975\\
 \hline
  \multicolumn{5}{|c|}{regBERT} \\ \hline 
  OSE & 0.929 & 0.706 & 0.651 & 0.561\\
 \hline
 Vikidia-En& 0.982 & 0.904 & 0.904 & 0.952\\
 \hline
\end{tabular}
\caption{Cross-Corpus Pairwise Ranking (Trained on Newsela-En)}
\label{tab:rankingcross}
\end{table}

NPRM model, trained on Newsela-En, does well with ranking both OSE and Vikidia-En texts by their reading level, and is more robust to variation among the corpora, compared to the regBERT model. All measures achieve performance $>0.87$ for both the datasets with \textit{$NPRM$}. The regBERT performs comparably on Vikidia-En, but does poorly on OSE. While the results for $NPRM$ are still somewhat lower in the cross-corpus evaluation than in within corpus evaluation setups, it has to be noted that this evaluation is done without any additional fine-tuning on the target datasets. We did not test for statistical significance in this case as the numbers have large differences between regBERT and $NPRM$ in most cases. This experiment leads us to a conclusion that $NPRM$ can successfully be used to rank documents on a different reading level scale too. 

\subsection{Zero shot, cross-lingual pair-wise ranking}
Zero-shot cross-lingual scenario aims to evaluate whether a model trained on one language can be effectively used to rank texts from another language correctly without explicitly training on the target language. We evaluated NPRM and regBERT models trained with a multilingual BERT (mBERT) model as the base for this task. Both the models were trained on Newsela-En dataset and evaluated on Newsela-Es and Vikidia-Fr datasets. The mBERT\footnote{\url{https://huggingface.co/bert-base-multilingual-uncased}} model is pre-trained on a corpus of multilingual data from 104 languages, including all the three languages in our experiment: English, French and Spanish.  Table ~\ref{tab:rankingcrossling} shows the results of this experiment. 

\begin{table}[htb]
\begin{tabular}{ |p{2.45cm}|p{0.85cm}|p{0.7cm}|p{0.8cm}| p{0.7cm}|}
 \hline 
 \multicolumn{5}{|c|}{$NPRM$(mBERT)} \\ \hline
 Testset & NDCG & SRC &KTCC & RA\\
 \hline
 NewsEla-Es& 0.996 & 0.985 & 0.971 & 0.864\\ 
 \hline
 Vikidia-Fr& 0.930 & 0.622 & 0.622 & 0.811\\
 \hline
  \multicolumn{5}{|c|}{regBERT (mBERT)} \\ \hline
  NewsEla-Es & 0.992 & 0.957 & 0.931 & 0.741\\
 \hline
 Vikidia-Fr& 0.913 & 0.527 & 0.527 & 0.764\\
 \hline
\end{tabular}
\caption{Zero-shot, cross-lingual Evaluation \\ (Trained on Newsela-En)}
\label{tab:rankingcrossling}
\end{table}

We observe that the $NPRM$ with mBERT either performs comparably or outperforms a regression mBERT model on all metrics, for both the datasets. Specifically, the $NPRM$ has a performance increase of $12.3\%$ in RA for Newsela-Es over Vikidia-Fr. Thus, we can conclude that our pairwise ranking approach performs well even in cross-lingual scenarios, and zero-shot, cross-lingual transfer can be useful to setup strong baseline models for new languages. 

We can notice a lower performance on Vikidia-Fr compared to Newsela-ES. Apart from the fact that they are different languages, it can potentially also be due to the fact that Newsela-ES has content from the same domain as Newsela-EN, whereas Vikidia-Fr has more diverse content.  It is also possible that the ranking metrics penalize Vikidia-Fr predictions more, as there are only two reading levels. A ranking can still be scored well if most of the ranking order is correct. However, in the case of Vikidia-Fr, an incorrect ranking levels would result in a completely reversed list, which is heavily penalized in SRC and KTCC.  Thus, small number of completely incorrectly ranked slugs can result in low SRC and KTCC scores for Vikidia-Fr, but can still result in high SRC and KTCC scores for NewsEla-ES. More future experiments, with additional languages, would lead us towards a better understanding of what works well across languages and datasets. 

\paragraph{Ranking Metrics}: We reported four ranking metrics in these experiments. While they all get high numbers in some experimental settings, none of them consistently seem like a better choice than others. We observe that the large majority of the methods score close to 1.0 on NDCG. In comparison, the SRC and KTCC, while generally quite high, appear more susceptible to poor ranking performance. We notice that RA is lower than SRC and KTCC for OSE (Table \ref{tab:rankingcross}) and NewsEla-Es (Table \ref{tab:rankingcross}), but SRC and KTCC lag behind RA for Vikidia-Fr (Table \ref{tab:rankingcrossling}). We hypothesize that this could be because of the number of reading levels in the datasets. SRC and KTCC seem more forgiving when number of reading levels are more. 

Clearly, each metric addresses the evaluation of ranking differently, and as the results show, there is no single model that consistently does well across all metrics, in all the evaluations. We hope that this illustrates the value of reporting multiple metrics while benchmarking a new model. Future work in this direction should also focus on the evaluation of the evaluation metrics themselves for this task.

\section{Conclusions and Discussion}
\label{sec:disc}
In this paper we proposed a neural pairwise ranking model for ARA ($NPRM$). We performed within corpus, cross-corpus and cross-lingual evaluations to benchmark $NPRM$. Our results in the context of the research questions we started with (Section 1) are discussed below:
\begin{enumerate}
 \item \textit{Is neural, pairwise ranking a better approach than classification or regression for ARA, to achieve cross-corpus compatibility?} - While regression, classification, and pairwise-ranking models all achieve comparable performance in a within-corpus scenario, pairwise ranking performs better in cross-corpus and cross-lingual evaluation scenarios.  
 
    \item \textit{Is zero-shot, cross-lingual transfer possible for ARA models through such a ranking approach?} - Our experiments show that zero-shot cross-lingual ARA is possible with pair-wise ranking. Our proposed model, $NPRM$, trained with English texts achieved $>80$\% ranking accuracy on both NewsEla-Es and Vikidia-Fr datasets. 
\end{enumerate}

\paragraph{Limitations of NPRM: } $NPRM$ models the relative reading difficulty level between texts.  While this approach has performed well for our generalizability experiments, there is a general lack of interpretability with $NPRM$. For example, the NewsEla-en dataset contains reading level designations that align with the common core educational standards \cite{Porter.McMaken.ea-11}, and are interpreted to match the school grades of U.S students from kindergarten to high school. Since the aggregation process of $NPRM$ sums the predicted scores between pairwise comparisons of an intended ranking, the aggregated score is bounded above by the input list size, is unlikely to correspond to the original reading level scale. Further, $NPRM$ takes a list of texts as input and the model forces the constraint of having at least two texts to be ranked as input. Hence, $NPRM$ is suitable only for scenarios where ranking by reading level is useful (e.g., ranking of search results by their reading level).

\paragraph{Outlook: }All the five datasets used in these experiments come primarily from news and wikipedia articles. However, the ARA is also studied in other domains (e.g., financial disclosures \cite{Loughran.McDonald-14}). Future work can test the validity of the conclusions of this paper in new domains. Further, all the datasets are human created texts. It would be interesting to explore how the model works for applications like text simplification and machine translation, which can support existing research on evaluating machine generated text. 

\section*{Ethics Statement}
In this paper, we report on the creation of a new dataset for readability assessment. The data collection process did not involve any human participants. So, no ethics board approval was necessary. Both the websites are available under permissive licenses that allow sharing and redistributing. The released dataset will follow the same procedures and is freely available at \url{https://zenodo.org/record/6327828}. An important point to note in the use of the dataset is that the length of texts is much shorter in the "simple" versions compared to regular Wikipedia articles, which may affect the quality of results in some use cases of the dataset. 

\section*{Acknowledgements} We thank the four anonymous reviewers for their helpful comments on the submitted manuscript. We also thank Gabriel Bernier-Colborne and Taraka Rama for their comments on the initial draft of this paper, and Michel Simard and Yunli Wang for the early discussions on this topic.

\bibliography{custom}
\bibliographystyle{acl_natbib}

\appendix

\end{document}